Preliminary results of a therapeutic lab for promoting autonomies in autistic children

Cristina Gena, Rossana Damiano, Claudio Mattutino, Alessandro Mazzei, Stefania Brighenti
Dipartimento di Informatica, Università di Torino, cristina.gena@unito.it

Matteo Nazzario , Valeria Ricci
Intesa Sanpaolo Innovation Center, corso Inghilterra 3, Turin, 10138, Italy

Camilla Quarato, Cesare Pecone, Giuseppe Piccinni
Jumple srl, Via Isonzo, 55/2, Casalecchio di Reno, Bologna, 40033, Italy

Federica Liscio, Loredana Mazzotta, Andrea Meirone, Francesco Petriglia
Fondazione PAIDEIA Onlus, Italy

This extended abstract describes the preliminary quantitative and qualitative results coming from a therapeutic laboratory focused on the use of the Pepper robot to promote autonomies and functional acquisitions in highly functioning (Asperger) children with autism. The laboratory[1] started in February 2021 and lasted until June 2021, and the weekly meetings lasted two hours and were led by one or two therapists (educators, speech therapists, psychologists, etc.), helped by 2-3 trainee master students. The participants recruited were four highly functioning (Asperger) children, aged between 11 and 13 years. There have been in total 16 lab sessions, all recorded by a fixed camera, in addition to the Pepper's 2D cameras. Furthermore, trainees filled out evaluation forms provided by psychotherapists, noting the children autonomy's progress in a diary with the helping of rating scales [1]. These notes were then reworked to draw up shared reports, reflecting on the behavior's evolution and progress of the children meeting by meeting.

The setting of the lab was an elegant apartment furnished as a real home in the city center. Each meeting had a similar structure: 1) welcome in the apartment; 2) social moment: dialogue with the robot; 3) moment of snack preparation; 4) moment of post-snack dialogue; 5) final feedback and goodbye.
The snack preparation was one of the most stimulating moments for the children, dedicated to the preparation, in the kitchen or directly on the dining room table, of some increasingly complex snacks. The group was led both by Pepper, instructed to organize and coordinate the activity, and by the therapists, ready to intervene when required.

The goal of the activity was to gradually mitigate the therapist's aid, so that therapists could only make suggestions from time to time. Pepper, with the help of the video–modeling [5] encouraged the participants to schematically organize themselves, listing the ingredients, illustrating the procedures with images, animations, and videos, and giving the children time to manage the preparation, as well as the possibility of reviewing the steps. The activity gave good results: the children appreciated the help of Pepper, showing good levels of increasing autonomy, even if, sometimes, the difficulties in maintaining a high concentration affected the scores reported by the trainees.
During the social moment, Pepper was conversing with children. On the one hand, the robot responded to their curiosity about itself, and, on the other, guiding a dialogue called "making friends", in which the robot attempted to establish a link with each participant, according to their previously declared interests, as advocated in the design of social and educational robots [2],[3], also target to autistic users [4]. Indeed, from time to time, Pepper re-proposed the topics children previously liked most. In the first case, the scenario envisaged that the children took their turns facing the robot, waiting for it to catch their gaze and listen, and then ask it any question. In the second case, the robot called the children one by one and began to talk with them on a previously liked topic (e.g., music, video games, etc.), which followed a script manually updated week by week, with the robot trying to guide the conversation. However, both the attempts led to unsatisfactory results, often arousing frustration among participants.

---

[1] Ethical approval for this study was obtained from the bioethical committee of the University of Turin, with approval number: 0664572

In fact, it could be argued that the problem arose at the roots of design, since both activities were very far from what we could really define a "conversation", other than a simple transmission of information. Bringing the mind back to the sociologist Sherry Turkle [6] conversations convey much more than the details of an argument: it is not just a question of answers, but of what they mean. As the developers did not implement real dialogue autonomy in the robot, Pepper showed no progress in the interaction, leaving the trainees to take note of the children's inclinations, and planning, from one meeting to the next, a new dialogue that considered what emerged.

The results from this experience showed some critical issues to be addressed in future works. Concerning the dialogue system, at least two related features need to be empowered. On the one side, the dialogue system needs to be improved in robustness and in precision. The actual conversations show a high degree of expectation from children about the robot's knowledge: to fulfill this expectation, one needs to have a correct and precise semantic representation of the children's questions in encyclopedic and commonsense domains. A possible improvement could be based on the construction of an annotated corpus by using a Wizard of Oz approach [7]. In this way, one could train a machine learning frame based natural language understanding system, starting from the annotation of user intents. The other improvement concerns the preparation of back-up dialogue strategies that the dialogue system can adopt in the case of non-comprehensible questions/utterances from the child or sentences not strictly related. An interesting possibility for building a back-up strategy is using large language-models as BERT [8].

We also defined an ontology for the possible topics of children interests whose classes and properties have been defined by extracting them from DBpedia[2]. As future work, we will integrate the robot's dialogue with this knowledge base to make the robot able to navigate the ontology and reason on it, thus enriching its dialogue strategies.

Focusing on the specific autistic children's features, we must notice that the autistic functioning distorts the essence of the conversation. The exchange of utterances does not produce the pleasure of sharing but is functional to obtaining something more concrete, such as searching for information. If the goal is reaching a typical conversation, the results could always be unsatisfactory in this context. At the same time, this calls for the collection of conversational data targeted at this specific group of interactants.

---

[2] https://www.dbpedia.org/